\documentclass[11pt,reqno]{amsart}
\usepackage[utf8]{inputenc}
\usepackage[T1]{fontenc}
\usepackage[english]{babel}
\usepackage{graphicx}
\usepackage{amsfonts,amsmath,amssymb,amsthm}
\usepackage[colorlinks=true, urlcolor=blue, linkcolor=blue, citecolor=blue, pdftex]{hyperref}
\usepackage{ulem}
\usepackage[a4paper,body={160mm,235mm},centering]{geometry}

\newtheorem{definition}{Definition}

\newtheorem{remark}[definition]{Remark}

 \usepackage{subcaption}
\usepackage{enumitem}
\begin{document}

\title[Explainable multi-class anomaly detection on functional data]{Explainable multi-class anomaly detection on functional data}

\author[M. Cura]{Mathieu Cura}
\address{Optimistik, 536 rue Costa de Beauregard, 73000 Chambéry, France}
\email{mathieu@optimistik.fr} 

\author[K. Firdová]{Katarína Firdová}
\address{Univ. Grenoble Alpes, Univ. Savoie Mont Blanc, CNRS, LAMA, 73000 Chambéry, France}
\address{Optimistik, 536 rue Costa de Beauregard, 73000 Chambéry, France}
\email{katarina.firdova@optimistik.fr}

\author[C. Labart]{Céline Labart}
\address{Univ. Grenoble Alpes, Univ. Savoie Mont Blanc, CNRS, LAMA, 73000 Chambéry, France}
\email{celine.labart@univ-smb.fr}

\author[A. Martel]{Arthur Martel}
\address{Optimistik, 536 rue Costa de Beauregard, 73000 Chambéry, France}
\email{arthur.martel@optimistik.fr} 

\date{\today}


\begin{abstract}
  In this paper we describe an approach for anomaly detection and its explainability in multivariate functional data. The anomaly detection procedure consists of transforming the series into a vector of features and using an Isolation forest algorithm. The explainable procedure is based on the computation of the SHAP coefficients and on the use of a supervised decision tree.  We apply it on simulated data to measure the performance of our method and on real data coming from industry.
\end{abstract} \hspace{10pt}

\maketitle


\section{Introduction}\label{Introduction}

\subsection{Anomaly detection in the industrial process}\label{Anomaly detection in the industrial process}
Digitalisation in the industry enables us to collect large amounts of data from various sources (sensors, MES, shift notebooks) and the analysis of such data has become critical to business operations such as system health monitoring, application performance monitoring or behavior analysis.\\

Anomaly detection in the industrial process is a large subject which requires a different approach for each use case. The main goal of anomaly detection is to find out the behavior which deviates from the typical one (later referred as a normal behavior) and identify its source in order to assess whether it requires an action. It allows a quick fixation of a rising problem and avoids or reduces breaks of the process.  \\

The definition of the anomaly itself is not unambiguous. While a particular behavior observed in data captured from a sensor can indicate a serious problem in case A, the same phenomenon can be observed as an expected reaction after change of operating mode in case B.

The tolerance of mistakes differs also. While in one domain it is crucial to detect all possible fails, an other domain, where each unnecessary shutdown caused by a false alarm is lossy, can be more willing to let minor anomalies unnoticed. \\

It is naive to expect a single algorithm which will point out real anomalies in all kinds of domains/process types and the client’s expert knowledge is necessary to specify the objectives. However, it is possible to construct a procedure with minimum individual modifications (e.g. choosing the most appropriate evaluation metric or the length of the period to be assessed) where the core principle remains the same and leads to the desired results.  \\

\subsection {Functional data}
\label{Functional data}
\begin{figure}[!ht]
\centering
\includegraphics[width=1\textwidth]{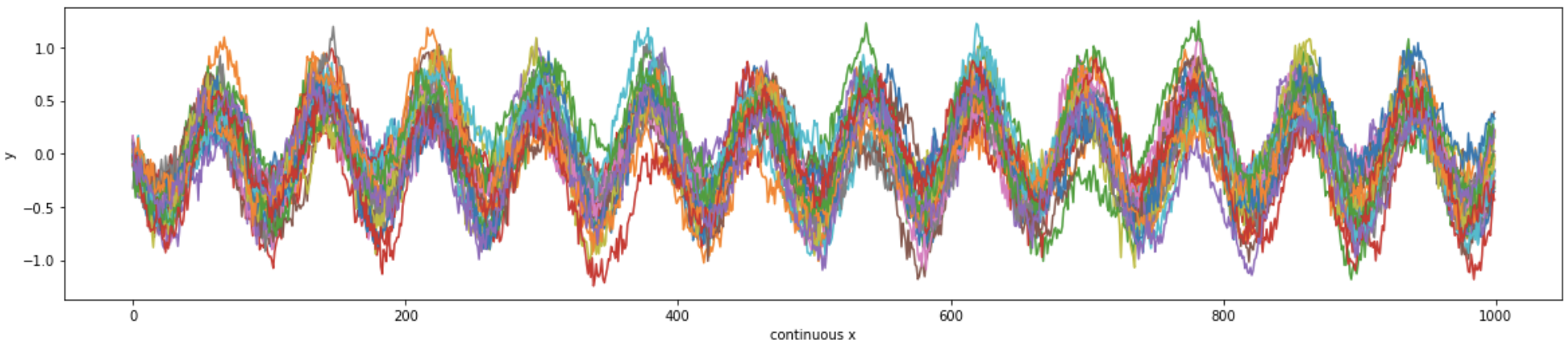}
\caption{Example of univariate functional dataset. Each observation is represented as a curve.}\label{func_data}
\end{figure}

In a functional dataset, each observation is a series of interdependent values which can be represented as a curve or a function (see example in Figure \ref{func_data}). 
\cite{FDAmuller} talks about data samples consisting of random functions or surfaces, where each function is viewed as one sample element. According to the definition given in \cite{FDAhooker}, functional data are multivariate data with an ordering on the dimensions. This kind of dataset is often found in practice since it can correspond to repeated measurements over some continuous domain, e.g. time or frequency. \\

In this paper we will focus on multidimensional functional data collected on a time domain, discretised for the numerical reasons. In a set of $n$ observations, the $i-$th observation for $i=1, \hdots ,n $ can be defined as $\{\mathbf{y}_t\}_{t\in \{1,T\}}$ where $\mathbf{y}_t$ is an $m-$dimensional vector and $T$ is the horizon time. Values in one series are interdependent, meaning that the value at time $t$ depends on the value at time $t-1$, while the observations themselves are independent of each other. \\

{ \it Example 1 :} If we define a certain time-and task-framed part of an industrial production as a batch, then each batch should represent the same action and should be described by the same function. When $n$ batches of duration $T$ are observed and $m$ different parameters are continuously recorded during each batch, creating $m-$dimensional series, we obtain a dataset of functional data. \\

{\it Example 2 :}  In case of a continuous production with repeating pattern, a continuous series can be divided into shorter segments. 
If this division is done taking into account the behavior of the process and if the segments are homogenous, each capturing the same phenomenon, (e.g. in a cyclic process each segment contains a full cycle), it also defines a functional dataset. \\
 
In the literature the subject of functional data is covered in \cite{Ferraty}, \cite{FDAmuller} or \cite{Ramsay}. A few papers deal with anomaly detection but mostly for univariate observations, e.g. in \cite{Febrero-bande} or \cite{FBoxplot}. \cite{lejeune} or \cite{FIF}  proposed an extension to the multivariate data but their methods are focused to detect specific kind of abnormal behavior and do not combine different types of anomalies.

\subsection {Anomaly detection on functional data}
\label{Anomaly detection on functional data}
As mentioned in the previous section, an observation in a functional dataset is a series. The observation can be identified as anomalous if it contains an anomaly of local or global type.

We define a local anomaly on a series as a short anomalous behavior, occurring during a limited period of the time domain, e.g. punctual rise of the noise. 
A global anomaly (or persistent anomaly, see \cite{hubert}) has an anomalous behavior on the whole time domain.\\ 

In the context of anomaly detection on functional data, the objective is to detect anomalous observations, i.e. multidimensional series containing anomalies.

In Example 1, data collected in batch production create a set of independent series, each series representing the behavior of a set of parameters during one batch. In this case an observation containing an anomaly points out the problematic batch. The problematic batch can contain a short change of behavior of one or more parameters (local anomaly) or can be different for the whole duration (global anomaly). An anomalous segment in Example 2 informs us about part of the continuous process during which an anomaly occurred. \\



\subsection{Explainability}\label{Explainability}
Without a good explanation, anomaly detection is only
of limited use : the end user knows that something anomalous has
just happened, but has little understanding of how it has arisen, how
to react to the situation, and how to avoid it in the future. The next step consists of finding the best explanation for the detected anomalies, or more precisely, a human readable formula offering useful information about what has led to
the anomaly.  
Explainability for time series can be done by using dimension reduction (see e.g. \cite{BSJD21}) or deep learning methods (see e.g. \cite{AGP20}). Recently, \cite{Exathlon} proposed a benchmark for explainable anomaly detection over time series data, pointing to the missing research in this domain. The paper compares methods for explainability of the anomalies found by neural network models, focusing on high-dimensional streaming processes (see EXstream \cite{exstream}, MacroBase \cite{macrobase} and LIME \cite{lime}). \cite{tree_Wu} used a tree-regularization approach to interpret deep time series models. \\

In this paper we present a method detecting multi-class anomalies in multivariate functional data and explaining these anomalies. In the first section we have described the subject in the context of industry. Functional data were introduced as a specific kind of industrial data. In the second section we describe our method which transforms a series into a vector of features, detects the anomalies and identifies the leading features of each group of anomalies. In the third section, we apply our methodology on two generated datasets and on a real one and present numerical results.  \\

\section{Methodology}\label{Methodology}

\begin{figure}[h]
\caption{Steps of the anomaly detection process mentioned in this section. From left to right: preparation of the functional dataset by collecting series corresponding to the batches or by dividing continuous process; extraction of the features and transformation of the series to the multidimensional vectors; detection of anomalous observations among vectors of features; grouping similar anomalous observations; identification of the features which contributes to the high score and characterise the anomaly type.}
\centering
\includegraphics[width=1\textwidth]{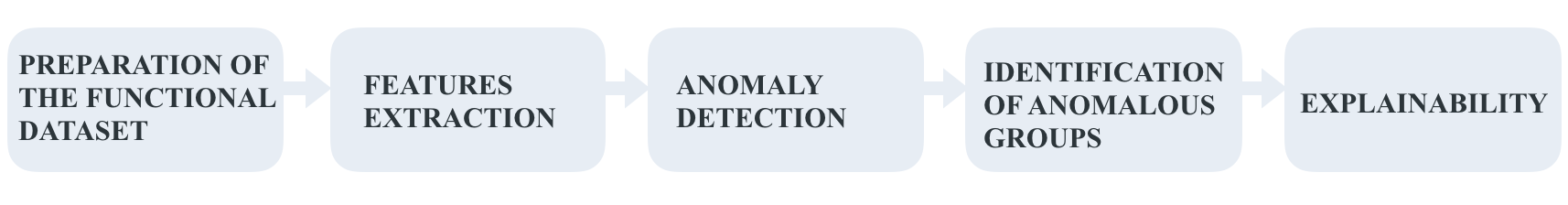}
\label{workflow}
\end{figure}

This section describes the process of anomaly detection starting with initial requirements, transformation, algorithm and explainability (workflow in Figure \ref{workflow}). Data preparation and preprocessing is not detailed in this paper. The proposed methodology is applicable in various situations which is a benefit in the industrial domain where each use case deals with different types of data.

\subsection{Data quality and limitations}
\label{Data quality and AD limitations}
We suppose that reader is familiar with basics of data preprocessing and knows that the data input should not contain any large period of missing values or permanent high noise, neither periods which we do not desire to evaluate, e.g. period of the shutdown. 
It is out of scope of this paper to provide the reader with full data collection procedure, but we emphasize that proper dataset is the base pillar for the future analysis and data quality will be reflected on the result of the scoring algorithm. 
Erroneous values create obstacles even for powerful algorithms.

\subsection {Series transformation and features extraction}
\label{Series transformation and features extraction}
In order to unify the process of anomaly detection as much as possible for various use cases we propose to use the approach which consists of transforming a series into multidimensional points and detecting anomalies.
The advantage of this approach relies on the fact that we do not use any assumption about data (e.g. we do not require the labels, non-stationary process,…) and it has adjustable steps hence it is suitable for different cases.  \\

The idea of the series transformation is to represent a time series as a vector of features characterizing the original series. We assume that all normal time series have similar behavior, therefore their features have similar values. On the contrary, time series with different behavior will differ in one or more of features. 

This transformation helps to reduce quantity of data which needs to be stocked and evaluated. Instead of considering each time instance individually, we can operate with multidimensional vector/point where each dimension corresponds to one feature. Moreover, vectors representing a time series are independent, which is not the case in a time series. This opens the door to many traditional anomaly detection algorithms. \\

An appropriate choice of handcrafted features allows to preserve information about the original behavior. Basic features (e.g. mean, standard deviation, maximum) can be completed by some advanced features (e.g. autocorrelation, skewness, power spectral density) using the knowledge of the process or information from the initial exploratory data analysis. A default group of features can be proposed to the user based on the character of the data (e.g. group of specific features if data are periodic).

Multivariate time series can be handled too. In a process where multiple parameters are recorded in parallel, features from each series can be extracted and concatenated into one vector. Information about all parameters are then held in one object and can be evaluated in once. If each parameter were evaluated separately, we would not be able to detect differences in relation of those parameters. \\

The method can be applied both in case of continuous production or in case of batch production. If the collected data are already in a form of the separated time series of the same length (Example 1), feature extraction is done on each time series independently. In case of continuous production (Example 2), the continuous series can be divided into shorter series and each series can be transformed individually. Note that this division should be done with regard to the behavior of the process and the series should be homogenous, each capturing the same phenomenon, e.g. in a cyclic process, a time series should contain a full cycle. As this paper deals with functional data, we do not cover the case when this division is not possible. A case when the batches do not have the same length is not covered neither. \\

The choice of the length of the segment used for the features extraction can be tricky. The series should be long enough to carry relevant information, yet a too long series can be a drawback in terms of response speed, i.e. if the anomaly occurs in the beginning of the segment, it cannot be seen before the end of the segment evaluation. This can be a problem if an anomaly occurs abruptly and leads to a fast deterioration of the industrial process, because we may not see it on time.

\subsection{Algorithm}
\label{Algorithm1}
As mentioned above, transformation avoids the problem of time-dependent points and represents a time series as a vector of many characterising features. The problem of anomaly detection in a time series is therefore converted into detection of anomalous multidimensional points. If the feature extraction has been done carefully, keeping relevant information about original behavior, the anomalous time series should deviate in one or more dimensions of its vector representation and we can apply an algorithm for anomaly detection in multivariate data. We propose to use Isolation forest \cite{IF} for its efficiency, speed and simplicity. \\

Contrary to methods which profile normal points (e.g. Inductive System Health Monitoring \cite{ishm}), Isolation forest is an algorithm which does not require any information about data distribution nor data labels, which make it suitable in many practical situations.
Computational complexity of Isolation forest grows linearly, making it preferable to methods like One-class Support Vector Machine \cite{ocsvm} or distance and density based algorithms (e.g. Orca \cite{orca} or Local Outlier Factor \cite{lof}).  \\

\begin{figure}[!ht]
  \caption{Example of possible repartitions of the data space from \cite{IF}.  Observation $x_0$ is isolated sooner than $x_i$.}
  \centering
  \includegraphics[width=.7\textwidth]{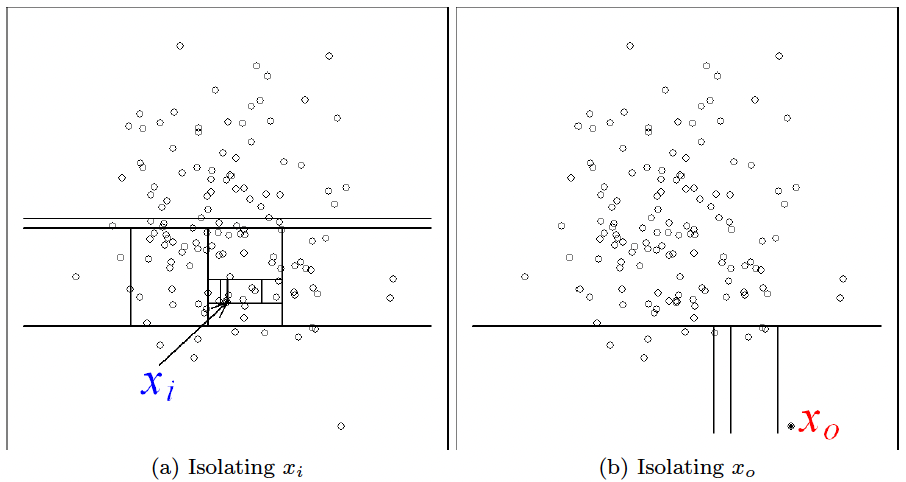}
  \label{IF-ex}
  \end{figure}

Isolation forest is an isolation based method which explicitly isolates anomalies. The idea of the method is that anomalies are few and different, hence they are isolated quicker - random partitioning produces shorter paths for anomalies, as illustrated in Figure \ref{IF-ex}. This simple structure encourages the choice of this algorithm, especially in practical applications where complicated algorithms may be viewed with skepticism from the user.  

The algorithm consists of the following steps:
\begin{enumerate}
	\item choose randomly a dimension and a random value between min and max values in that dimension
	\item split data by drawing a straight line (or hyperplane) through that value
	\item repeat 1.-2. until a tree is complete (with possibility to adjust size of a sample used for the creation of a tree)
	\item repeat 1.-3. until having a forest of multiple trees (with possibility to adjust number of trees in a forest)
	\item compute anomaly score for observation $x$ by averaging its path length from all trees
\end{enumerate}

Random partitioning produces significantly shorter paths for anomalies since instances with distinguishable feature values are more likely to be separated in early partitioning. Path length $h(x)$ of an observation $x$ is measured by the number of edges that $x$ traverses in a tree from the root node until it terminates at an external node. According to \cite{IF}, average path length converges well before the number of trees $T = 100$. Anomaly score of an observation $x$ is based on this length.


\subsubsection*{Threshold}
In practice we may know from the recorded events or clients feedbacks what part of the production is anomalous and use this information to choose the value of the anomaly score which has to be exceeded in order that an alarm is triggered. However, the exact value is not known and other aspects of the process should be kept in mind while adjusting the threshold.
In case of a lower threshold, it is more likely that all undesirable events are captured, but the drawback is an increased number of false alarms which can lead to losses due to a stopped production. Depending on the domain, the tolerance for faults varies as well as the cost of the shutdowns and a trade-off between detecting all anomalies and avoiding false alarms should be found. 

\subsection{Explainability}
\label{Explainability1}

More and more importance is placed on retrograde explainability of the results. From the practical point of view, the score itself is insufficient and it is important to know which parameters have contributed the most to the given score.  \\

One of the objectives can be to analyse groups of observations with the same type of anomaly and find which features define abnormality of those groups. This can be done in a two steps analysis: 
\begin{enumerate}
	\item Use a clustering algorithm to find similar clusters in the dataset (vectors of the features) and assign them cluster labels
	\item For each cluster, find the features which distinguish the given cluster from the others
\end{enumerate}

\begin{figure}[!ht]
  \begin{subfigure}[t]{.3\textwidth} 
    \centering
    \includegraphics[width=1.0\linewidth]{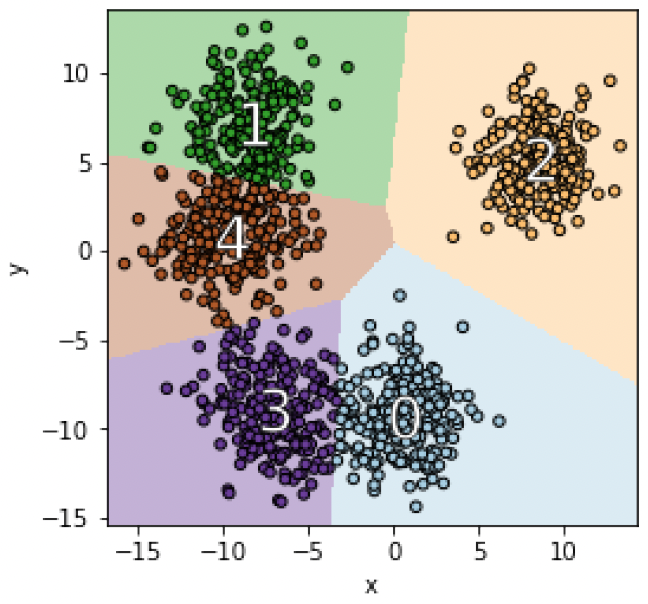}  
    \caption{Five clusters found by K-means in two-dimensional dataset}
    \label{exkmc1}
  \end{subfigure}
  \begin{subfigure}[t]{.3\textwidth}
    \centering
    \includegraphics[width=1.0\linewidth]{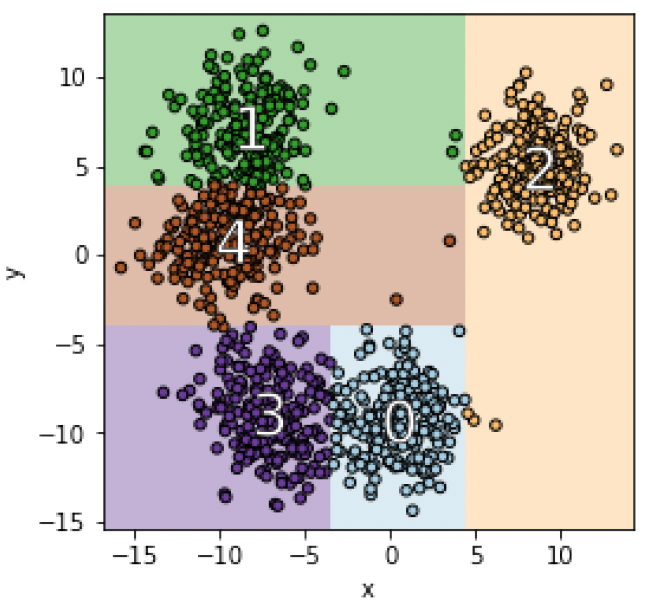}  
    \caption{Explainable clustering divides the space by straight cuts, trying to separate the clusters}\label{exkmc2}
  \end{subfigure}
  \begin{subfigure}[t]{.3\textwidth}
    \centering
    \includegraphics[width=1.0\linewidth]{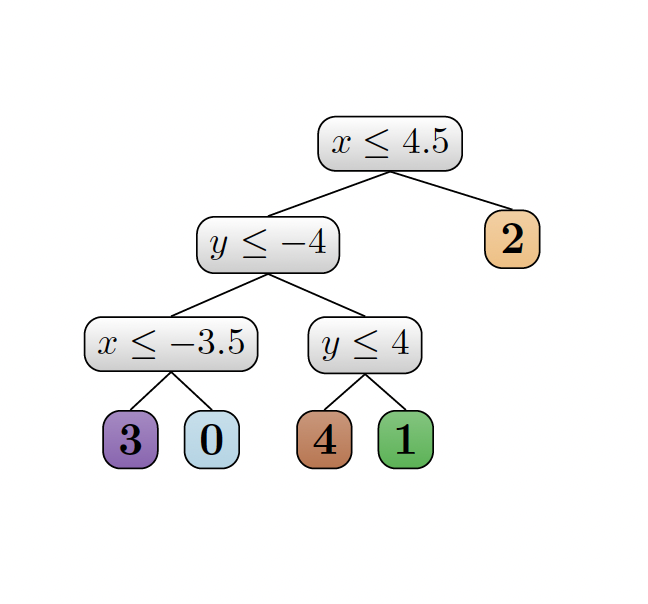}  
    \caption{Tree-like structure with thresholds and dimensions used to divide the space}\label{exkmc3}
  \end{subfigure}
  \caption{Principal idea of Explainable k-Means and k-Medians Clustering \cite{exkmc}}\label{exkmc}
  \end{figure}

In \cite{exkmc}, the authors unify these two steps and propose a method which use K-means to create clusters and the Iterative Mistake Minimisation algorithm to create tree-like structure which separates the found centroids, trying to minimise misclassification. An example of this process is presented in Figure \ref{exkmc}.
  

To add more possibilities in customisation we propose to carry out these two steps separately. The rest of this subsection will be focused on the second step, assuming that the first step can be done with traditional clustering algorithm like K-means. \\

To analyse which features are the most contributing ones to the anomaly scores of the observations belonging to the same group, tools like SHAP \cite{shap} can be used. SHAP (SHapley Additive exPlanation) have been recently developed in order to simplify understanding of machine learning models. They allow to explain the effect of each used feature on the final score for each observation. The sum of SHAP values for one observation is the difference between the final score $f(x)$ and the base value $E[f(x)]$ which would be predicted if we did not know any features. 

SHAP is based on Shapley values known from the game theory, where the idea is to look at the differences in the predictions of the model with and without a given feature, considering all possible subsets from the set of all features.

Hence, for the $i-$th feature the Shapley value $\phi_i$ is
\begin{equation}
\phi_i = \sum_{ S \subseteq F \backslash \{ i \}}  \frac{ |S|!( |F| - |S| -1)!}{ |F| !} [ f_{S \cup \{ i \} } (x_{S \cup \{ i \} }) - f_S(x_S)]
\end{equation}
where $F$ is the set of all features, $S$ is the subset of features, $f_{S \cup \{ i \} } (x_{S \cup \{ i \}})$ is the prediction of the model including the $i-$th feature and $f_S(x_S)$ is the prediction without the $i-$th feature.

To see how SHAP has incorporated this concept we refer the reader to \cite{shap}.  \\

Alternatively to the group analysis with SHAP, a traditional decision tree classifier can be used to find the most important features. Cluster labels assigned in the first step can be used to train the tree and each final class in the tree can be defined by the features leading to the creation of that class.


\begin{remark} SHAP values can also be used for clustering (Step 1).
The output of SHAP applied on the Isolation forest model are values which indicate the contribution of each feature to the Isolation forest anomaly score. SHAP values rescale the data and highlight the differences between the observations. We can therefore consider an execution of the clustering algorithm directly on these values. We compare both methods (clustering on standardized data and clustering on SHAP output) on simulated data in Section \ref{ExplainabilityD1}.
\end{remark}

\subsection{Evaluation}
\label{Evaluation1}
Parts of the process described in Sections \ref{Algorithm1} and \ref{Explainability1} can be evaluated individually on known datasets. \\ 

Quality of scoring with the Isolation forest algorithm can be verified with metrics like AUC/ROC curve. If we know the ratio of the anomalous observations in the dataset, we can simulate the alert.\\ 

The evaluation of the clustering algorithm is more complicated since most of the evaluation metrics assume that the classification is done on datasets with two possible outputs. In this paper, we evaluate clustering algorithms with weighted $F_\beta$ score (Equation \ref{equation2}), which can deal with imbalanced multiple classes. 

\begin{equation}
\label{equation2}
F_{\beta}= \frac{(1+\beta)^2 \times \mbox{Precision} \times \mbox{Recall} } {\beta^2 \times \mbox{Precision} + \mbox{Recall}}
\end{equation}


Choosing a low value for $\beta$ assumes that {\it false positives} are more costly while choosing a high value for $\beta$ assumes that {\it false negatives} are more costly. $\beta$=1 assumes equally costly misclassifications. 

However, the problem of the non-equality of the classes still remains. In order to consider the fact that dataset contains one normal class and possibly many anomalous classes, the evaluation is done in two separated steps:

\begin{enumerate}
\item Merge all anomaly clusters and compute confusion matrix for normal cluster vs all other anomaly clusters in order to evaluate the capacity of the clustering algorithm to separate the cluster with normal observations from the anomalies
\item Consider only anomalous clusters and compute confusion matrix among them in order to evaluate the capacity of the clustering algorithm to distinguish different types of the anomalies
\end{enumerate}


\section {Experimental results}
\label{Experimental results}

In this section we demonstrate the process of anomaly detection on three datasets. In order to cover various situations that may arise and evaluate the approach we use generated data which are supposed to imitate the real behavior of a process. In addition, one real dataset is also available. 

We will try to reproduce a common situation, i.e. when the labeled history of the data is not available and therefore we do not know how the normal observation neither anomalies look like.

\subsection{Data}
\label{Data}

Dataset n°1 has been generated in form of a periodical, two-dimensional time series $(X_1,X_2)$. For example, one observation could correspond to data from one batch where two parameters have been recorded. The first parameter has been generated as a sine function and the second parameter as a combination of sine and cosine functions, both with low-amplitude white noise. Five abnormal behaviors have been generated:
\begin{itemize}
    \item   Local noise with higher amplitude (noise during $300$ time-steps)
    \item   White noise throughout observation
    \item   Constant value during full observation
    \item   Different parameters in the generated function (normal function combined with cosine function during 200 time-steps) for a short period of time
    \item   Different parameters in the generated function during full observation (longer period)
\end{itemize}

\begin{figure}[ht]
  \begin{minipage}[b]{.46\linewidth}
  \centering \includegraphics[width=1\linewidth]{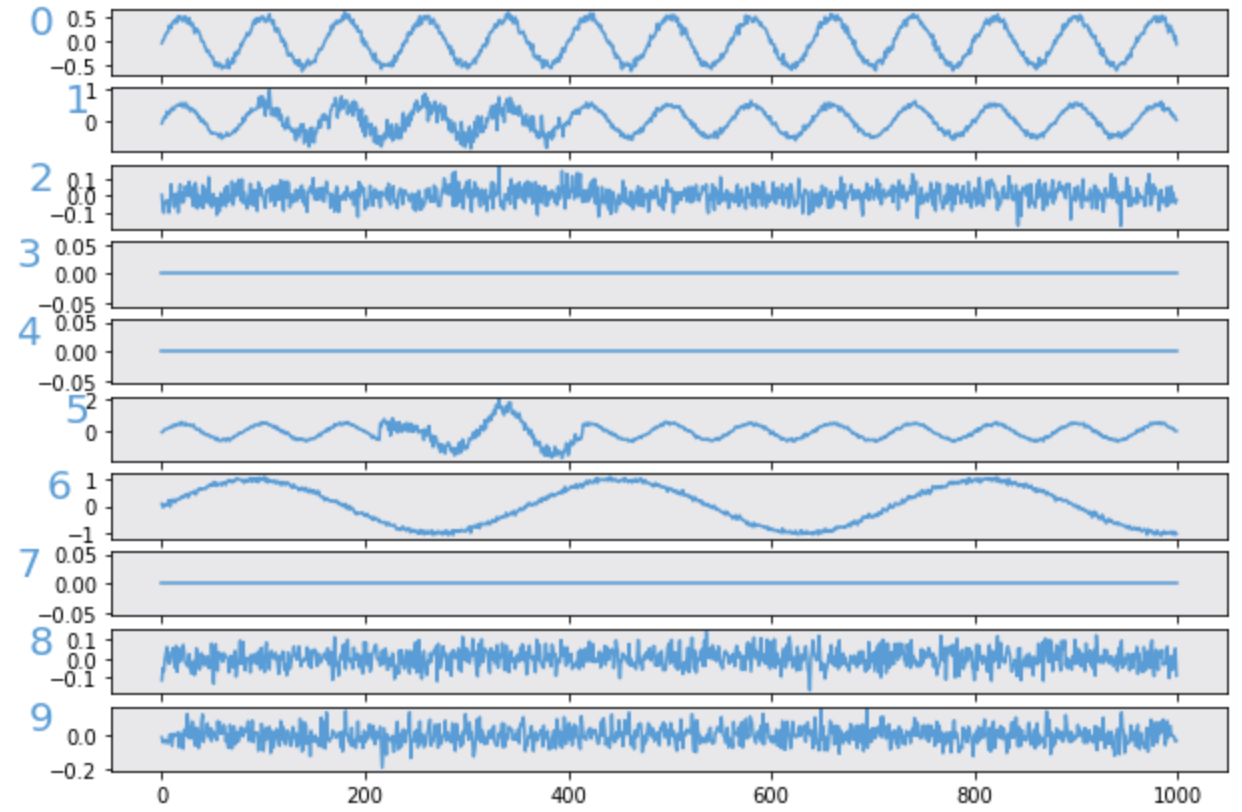}
  \title{$X_1$}
  \end{minipage} \hfill
  \begin{minipage}[b]{.46\linewidth}
  \centering \includegraphics[width=1\linewidth]{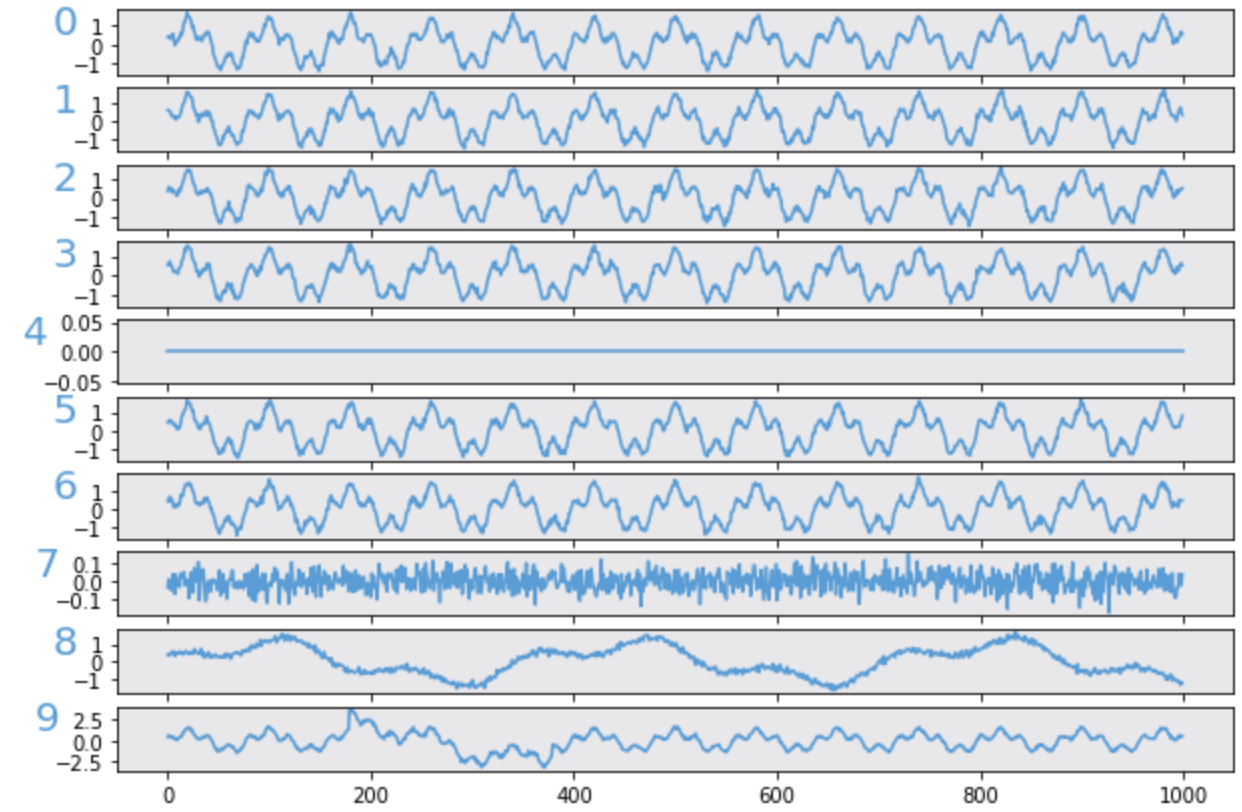} 
  \title{$X_2$}
  \end{minipage}
  \caption{The first dataset contains $N=500$ of $m=2$ dimensional functional series of length $l=1000$.
450 series of this dataset have normal behavior. Normal behavior is such that $X_1(t)= 0.5\sin(\frac{\pi t}{40}) + Y(t),  \; Y(t) \sim \mathcal{N}(0,0.5) $ and $X_2(t)=\sin(\frac{\pi t}{40}) + 0.5 * \cos(\frac{\pi t}{10})  +Y(t),  \; Y(t) \sim \mathcal{N}(0,0.1)$. $X_1$ is represented on the left hand side and $X_2$ is represented on the right hand side of the graphs. Line $0$ corresponds to one series with normal behavior, whereas 
 lines $1$ to $9$ correspond to different anomalous observations. 
 There are 50 anomalous series in the dataset. Each anomalous series is composed by one anomaly on $X_1$ (chosen in the above list) and/or one anomaly on $X_2$ (chosen in the above list).}
\label{data_per}
  \end{figure}

These abnormal behaviors have been combined and applied on a few series resulting in 9 different types of anomalous observations (Figure \ref{data_per}).
The objective is to detect those observations, whose behavior deviates in one or more parameters. In practice this corresponds to identifying anomalous batches. \\

\begin{figure}[!ht]
  \begin{minipage}[b]{.46\linewidth}
  \centering \includegraphics[width=1\linewidth]{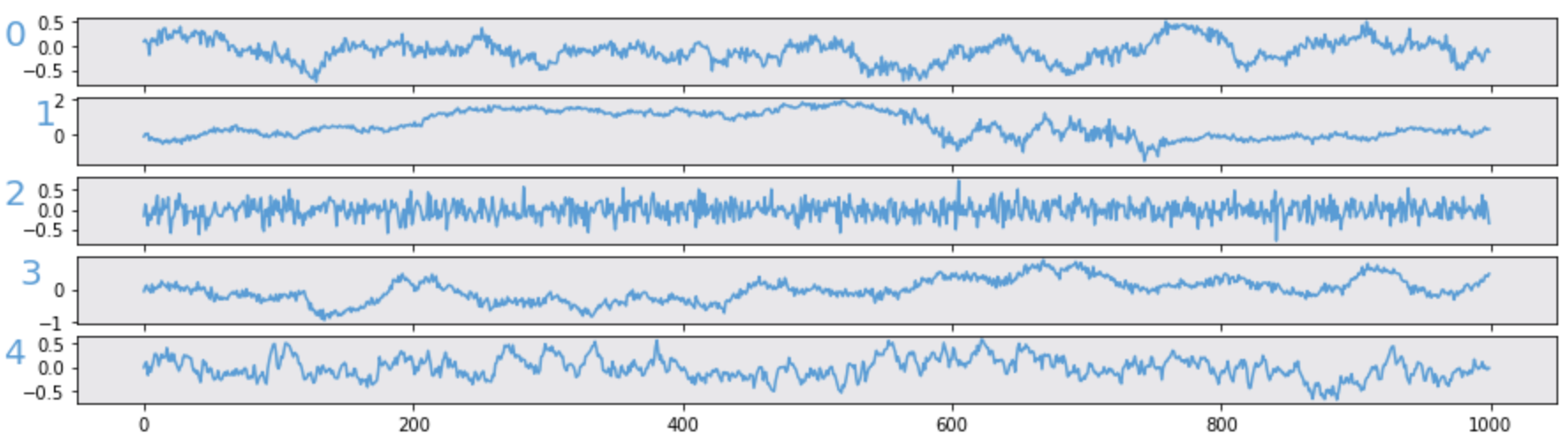}
  \title{$X_1$}
  \end{minipage} \hfill
  \begin{minipage}[b]{.46\linewidth}
  \centering \includegraphics[width=1\linewidth]{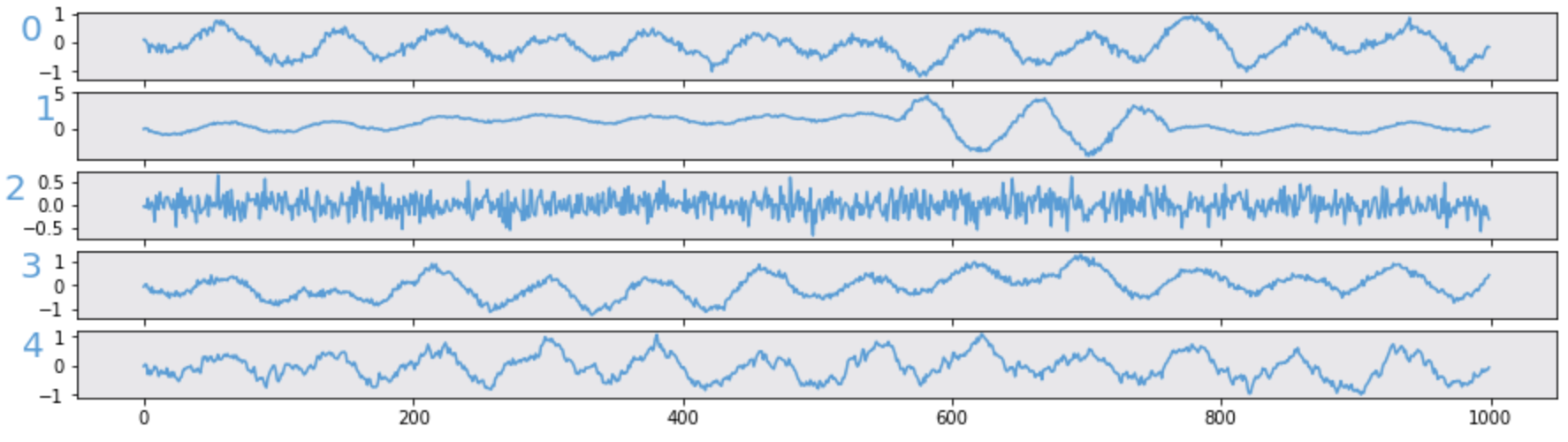} 
  \title{$X_2$}
  \end{minipage}
  \caption{The second dataset contains $N=520$ of $m=2$ dimensional functional series of length $l=1000$.
  450 series of this dataset have a normal behavior. Normal behavior is defined by $X_1(t)=0.4X_1(t-1)+0.3X_1(t-2)+0.2X_1(t-3)+0.05X_1(t-4)$ and $X_2(t)=X_1(t)+0.5\sin(\frac{\pi t}{40})$. Example of the normal observation is in the first lines of the figure. Line $0$ corresponds to one series with normal behavior, whereas lines 1 to 4 correspond to different anomalous observations. There are $70$ anomalous series in the dataset.}
\label{data_ar4}
  \end{figure}

{\it Dataset 2} (Figure \ref{data_ar4}) is of the same type as {\it Dataset 1}, i.e. one observation corresponds to a two-dimensional time series. The first parameter of the normal observation has been generated as an autoregressive process with lag 4. The second parameter is made up as an autoregressive process with lag 4 plus with sine function. The anomalies (occurring for both parameters) included in the dataset are:

\begin{itemize}
    \item  Local noise with higher amplitude (noise during 300 time-steps)
    \item   White noise during full observation
    \item   Change of the coefficients of lags 3 and 4
    \item   Coefficients of lags 2, 3 and 4 are set to 0, hence the resulting function is an AR(1) process
\end{itemize}

\begin{figure}[!ht]
\caption{Example of one observation from the third dataset. Original values of two measured parameters during a time interval of a few seconds. Anomalies are unknown.}
\centering
\includegraphics[width=0.8\textwidth]{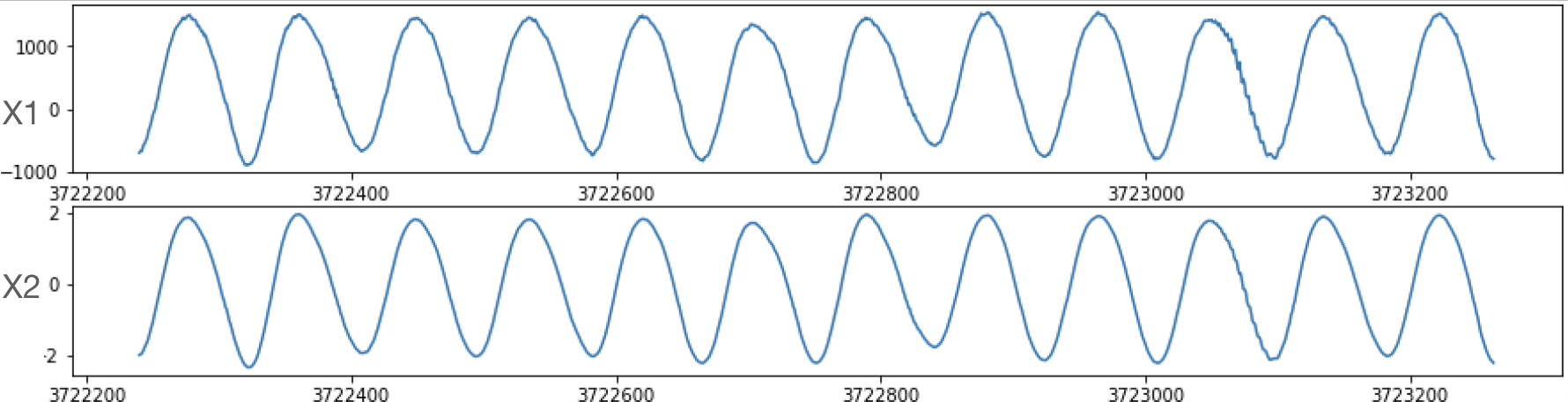}
\label{real_original_ts}
\end{figure}

{\it Dataset 3} (Figure \ref{real_original_ts}) is derived from real equipment data. Available data consist of sequence of snapshots taken regularly. 
Each snapshot records two parameters, $X_1$ and $X_2$. \\

This is an example of a continuous process where one time series correspond to a time-limited part (few seconds in this case) of the whole process and the objective is to identify which part of the process has a non usual behavior. Since it is a real example without historical database, we do not know which series are considered as anomalous from the point of view of the experts and we cannot evaluate the efficiency of our approach.

\subsection{Feature extraction}
\label{Feature extraction}

It is challenging to select all relevant features which should characterise the series if we do not know the dataset. Python package \texttt{tsfresh} is a tool which automatically computes a large number of time series characteristics in order to capture various aspects of behavior of the original series, but including all of them may lead to problems known as curse of dimensionality. \\

We can use information from initial visualisation as described in Section \ref{Series transformation and features extraction}. It is visible that data in {\it Dataset 1} and feature $X_2$ from {\it Dataset 2} have periodical character and in order to describe their behavior precisely it is important to include information about waves which compose the data, using tools as Wavelet Transformation, Fast Fourier Transformation or Power Spectral Density. 
Methods of signal processing are used to find the frequencies where the most of the signal is concentrated and help to choose the most relevant coefficients. \\

The choice of the features can also be motivated by knowledge of possible anomalies and level of distinction between them in the explanatory step. For example, if we want to distinguish a constant observation or a white noise from the periodic data, we may include information about autocorrelation of the series.

Among features proposed by \texttt{tsfresh} we have selected 15 features for {\it Dataset 1} and 13 features for {\it Dataset 2}, which will be visible in Figures \ref{ar4_shap_boxplot} and \ref{per_shap}.

\subsection{Generated data: Model and evaluation}
\label{Model and evaluation}

\subsubsection{Anomaly detection with Isolation forest}
\label{modelIF}
After the feature extraction step applied to each series of the dataset, we use the vectors created from these features as an input to the algorithm. As these vectors contain some information about original data, they should change in certain dimensions (one dimension corresponds to one feature) when the behavior changes, as described in the section \ref{Algorithm1}. \\

We randomly choose $70\%$ of the observations to train the algorithm. Consequently we use the trained model to assign anomaly score to each observation in the dataset.  \\

\begin{figure}[!htt]
\begin{subfigure}[t]{.45\textwidth} 
  \centering
  \includegraphics[width=0.9\linewidth]{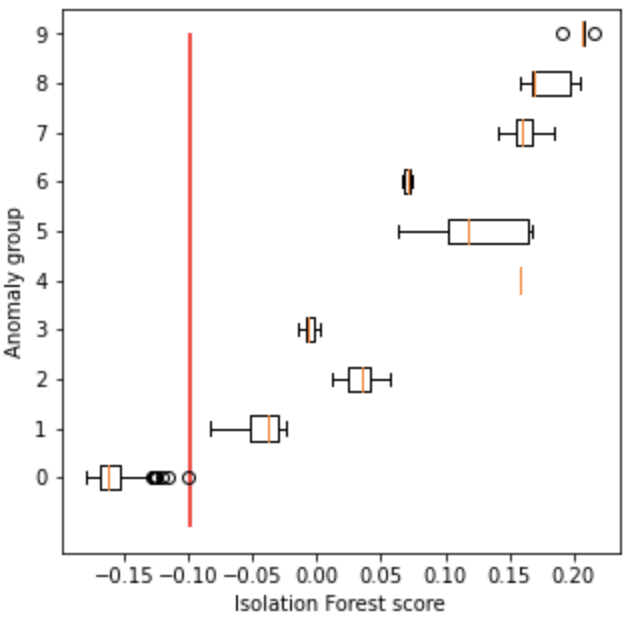}  
  \caption{ \it{Dataset 1}. Threshold = quantile $90\%$ . Area Under the Receiver Operating Characteristic Curve (ROC AUC) score = 1 }
\label{fig_if_per}
\end{subfigure} \hfill
\begin{subfigure}[t]{.45\textwidth}
  \centering
  \includegraphics[width=0.9\linewidth]{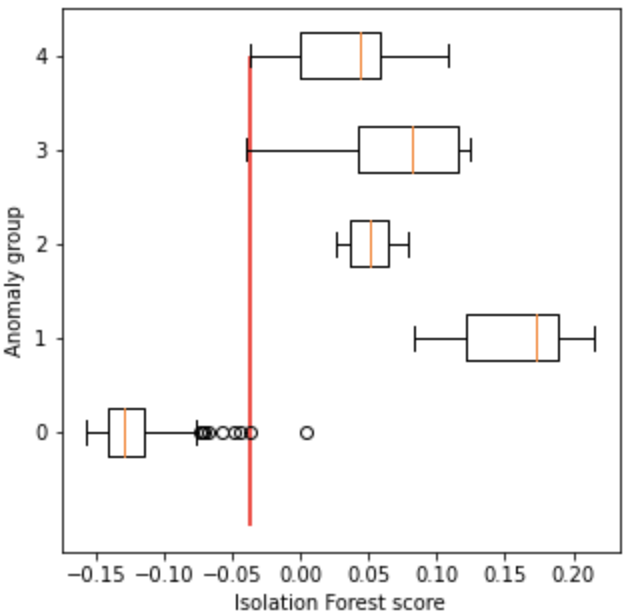}  
  \caption{  \it{Dataset 2}. Threshold = quantile $86\%$, corresponding to the ratio of anomalous observations in the dataset ($70$ of $520$).  Area Under the Receiver Operating Characteristic Curve (ROC AUC) score = 0.99}
  \label{IF_ar40_podiel}
\end{subfigure}

\caption{Output of the Isolation forest algorithm applied to the \it{Datasets} 1 and 2. Figures display the spread of the anomaly score (boxplots in $x$-axis) for each group ($y-$axis). Note that this division by group as well as the threshold are added to highlight the difference between low score of normal observations (group 0) and higher score for other groups. This information is not known in reality.}
\label{IF}
\end{figure}

Since in our generated datasets we know exactly which observations are abnormal, we can display anomaly score assigned by Isolation Forest for each anomaly group (see Figures \ref{fig_if_per} and \ref{IF_ar40_podiel}). Quality of scoring can be verified with metrics like AUC/ROC curve.   \\

\subsubsection{Explainability in Dataset 1}
\label{ExplainabilityD1}

We use the approach mentioned in Section \ref{Explainability1} to explain the different types of anomalies contained in the dataset. 


Clustering is done after the features extraction step and the transformation of the series. Since the tested clustering algorithms are based on Euclidean distance, input data should have the same range of values in all dimensions. We test clustering on standardised vectors of features and on the SHAP explanatory tool applied to the Isolation forest model. \\

Since the dataset contains imbalanced and unequal classes, the evaluation is done in two steps, as described in \ref{Evaluation1}. Results are presented in tables with the following columns:
\begin{itemize}[label={}]
 \item C1: Evaluation of normal vs all anormal observations, on SHAP output
 \item C2: Evaluation within anomalies,  on SHAP output
 \item C3: Evaluation of normal vs all anormal observations, on standardised data after feature extraction
 \item C4: Evaluation within anomalies, on standardised data after feature extraction
\end{itemize}

$F_\beta$ score for K-means on SHAP and on standardised dataset (after feature extraction) is given in Table \ref{tab1}.

\begin{table}[!ht]
  \caption{$F_{\beta}$ score for K-means \label{tab1}}
\begin{center}
\begin{tabular}{|c||c|c|c|c|}
  \hline
 - & C1& C2 & C3 & C4 \\ 
 \hline
 $\beta=0.5$ & 0.99 & 0.95 & 0.99 & 0.95 \\  
 \hline
 $\beta=1$ & 0.99 & 0.94 & 0.99 & 0.93 \\ 
 \hline
$\beta=2 $ & 0.99 & 0.93 & 0.99 & 0.92\\
\hline
\end{tabular}
\end{center}
\end{table}

$F_\beta$ score for BIRCH (Balanced Iterative Reducing and Clustering using Hierarchies) clustering algorithm on SHAP and on standardised dataset (after feature extraction) is given in Table \ref{tab2}.
\begin{table}[!ht]
  \caption{$F_{\beta}$ score for BIRCH \label{tab2}}
\begin{center}
\begin{tabular}{|c||c|c|c|c|}
  \hline
 - & C1& C2 & C3 & C4 \\ 
 \hline
 $\beta=0.5 $ & 0.99 & 0.50 & 0.99 & 0.98\\ 
 \hline 
 $\beta=1$ & 0.99 & 0.49 & 0.99 & 0.97\\ 
 \hline
$\beta=2 $ & 0.99 & 0.51 & 0.99 & 0.96\\
\hline
\end{tabular}
\end{center}
\end{table}

$F_\beta$ score for K-Medoids clustering algorithm on SHAP and on standardised dataset (after feature extraction) is given in Table \ref{tab3}.

\begin{table}[!ht]
  \caption{$F_{\beta}$ score for K-Medoids \label{tab3}}
\begin{center}
\begin{tabular}{|c||c|c|c|c|}
  \hline
 - & C1& C2 & C3 & C4 \\ 
 \hline
 $\beta=0.5 $ & 0.62 & 0.44 & 0.58 & 0.56\\  
 \hline
 $\beta=1$ & 0.44 & 0.42  & 0.40 & 0.59\\ 
 \hline
$\beta=2 $ & 0.37 & 0.45 & 0.33 & 0.64\\
\hline
\end{tabular}
\end{center}
\end{table}

As seen above, a modification of the parameter $\beta$ does not have any significant effect on the score. We have decided to use K-means algorithm because of its efficiency and popularity. Clustering will be done on SHAP output. \\



\subsubsection*{Analysis of the important features by SHAP}
\label{Important features}

\begin{figure}[!htt]
\caption{This figure shows importance of the features in each cluster in \it{Dataset 2}. Assuming that clustering algorithm has efficiently detected the five groups (A-E), we can identify the most important features ($y$-axis) based on their SHAP ($x$-axis). In group E, which corresponds to the normal observations, all features are of equal importance. We can discern groups where all important features within a group are similar, e.g. partial autocorrelation with different lags, or power spectral density with different coefficients, indicating the origin of the anomaly in the given group.} 
\centering
\includegraphics[width=1\textwidth]{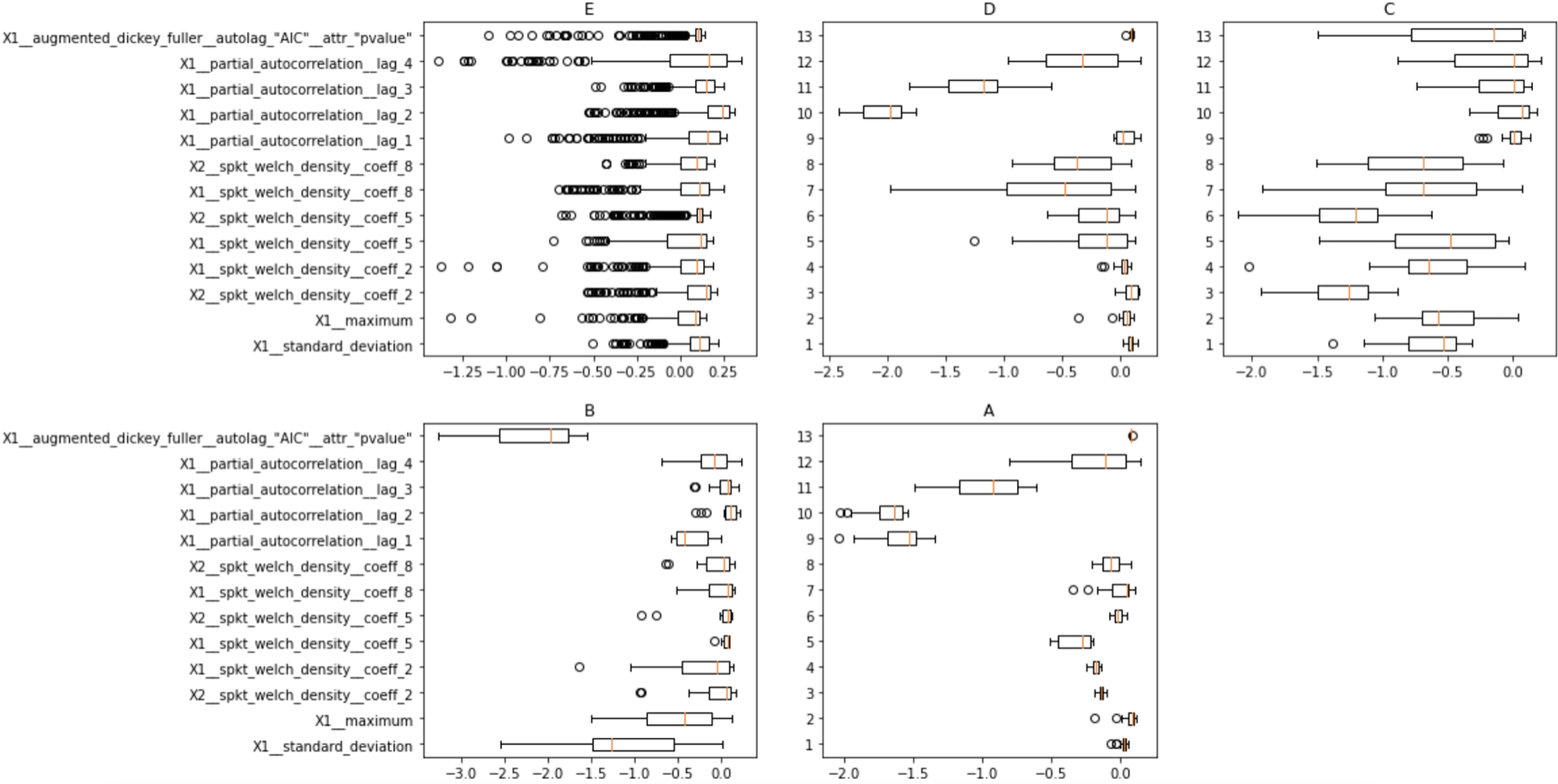}
\label{ar4_shap_boxplot}
\end{figure}

To analyse the detected anomaly types, we examine the SHAP values for each group. A high negative SHAP value of a feature in a group of observations indicates that the given feature has a significant decreasing effect on the score of this group, as demonstrated in Figure \ref{ar4_shap_boxplot}. \\ 

Figure \ref{ar4_shap_boxplot} displays different groups of observations (A-E) found by clustering algorithm. The number of groups is found by using the silhouette score. The groups correspond to the following anomaly types:
\begin{itemize}[label={}]
 \item E: Normal observations
 \item D: Coefficients of lags 2, 3 and 4 are set to 0, hence the resulting function is an AR(1) process
 \item C: Local noise with higher amplitude (noise during 300 time-steps)
 \item B: Change of the coefficients of lags 3 and 4
 \item A: White noise during full observation
\end{itemize}

It seems coherent that:
\begin{itemize}[label={}]
  \item In group D, features lag 2 and lag 3 partial autocorrelations are the most important. This is coherent since this anomaly has been generated by removing these coefficients;
  \item In group C, features related to the power spectral density indicate a change of the behavior;
  \item In group B, feature testing the stationarity of the series (Dickey-Fuller test) is the most important;
  \item In group A, features lag 1, 2 and lag 3 partial autocorrelations are the most important. This is also coherent since this anomaly is a white noise.
\end{itemize}

\subsubsection*{Analysis of the most important features by using a Decision tree}
\label{Decision tree}

\begin{figure}[!ht]
\caption{This figure represents an example of a decision tree on {\it Dataset 2}. In the first line of the decision node we can see which feature (the indicated number $i$ corresponds to the $i$-th feature listed in Figure \ref{ar4_shap_boxplot}) has been used to create clusters in the dataset. For example, to set aside the group of normal observations, it has used higher lag 3 partial autocorrelation (feature n°11) and lower standard deviation (feature n°1).} 
\centering
\includegraphics[width=0.8\textwidth]{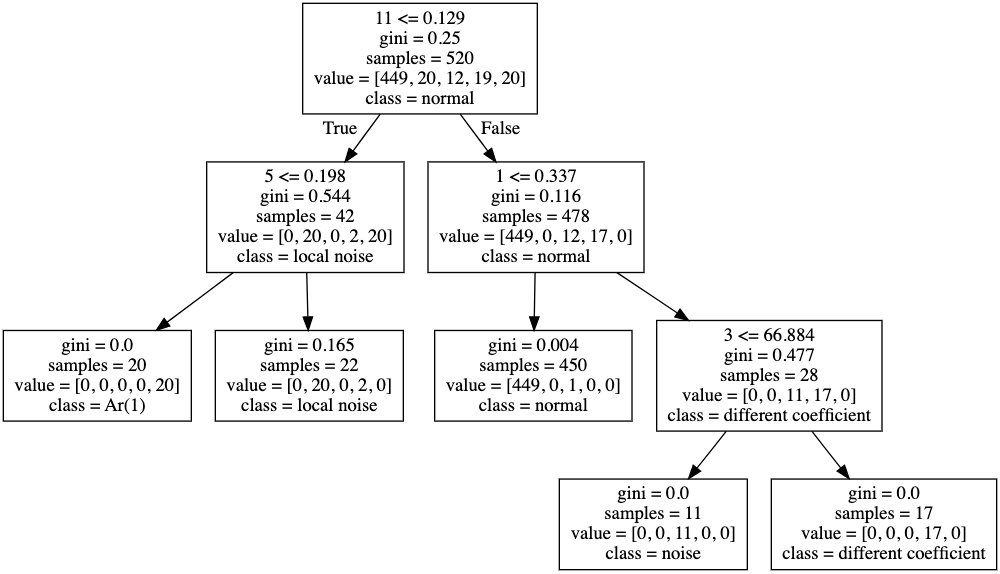}
\label{ar4_tree}
\end{figure}

The approach using a decision tree may be more comprehensible in practice as the concept of decision tree is easier to grasp. We use the labels given by the clustering algorithm to train the decision tree assuming that it will use relevant features in its condition nodes to create final leafs containing the same class.
Figure \ref{ar4_tree} represents an example of a decision tree. The following acronyms are used in the code to designate the different groups :

\begin{itemize}[label={}]
  \item \texttt{normal} for group E;
  \item \texttt{Ar(1)} for group D;
  \item \texttt{local noise} for group C;
  \item  \texttt{different coefficient} for group B;
   \item \texttt{noise} for group A.
\end{itemize}

\begin{figure}[!ht]
\caption{Overview of features importance ($y$-axis) found by decision tree. The criterion is Gini importance ($x$-axis), i.e. reduction of the split impurity brought by that feature} 
\centering
\includegraphics[width=0.7\textwidth]{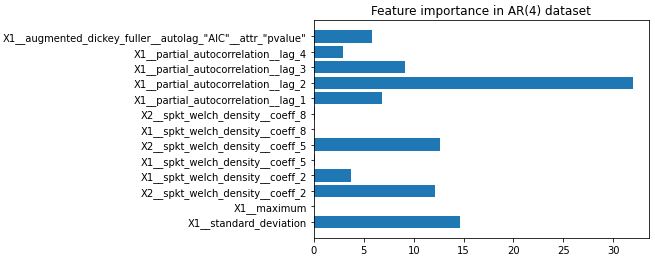}
\label{ar4_tree_features}
\end{figure}

The process has been repeated $100$ times in order to have more robust results. Impurity-based feature importances can be listed for the decision tree (Figure \ref{ar4_tree_features}), however it does not inform us which features are determining for a given class. \\

\subsubsection{Explainability in Dataset 2}
\label{ExplainabilityD2}

\begin{figure}[!ht]
\begin{subfigure}[t]{.3\textwidth} 
  \centering
  \includegraphics[width=\linewidth]{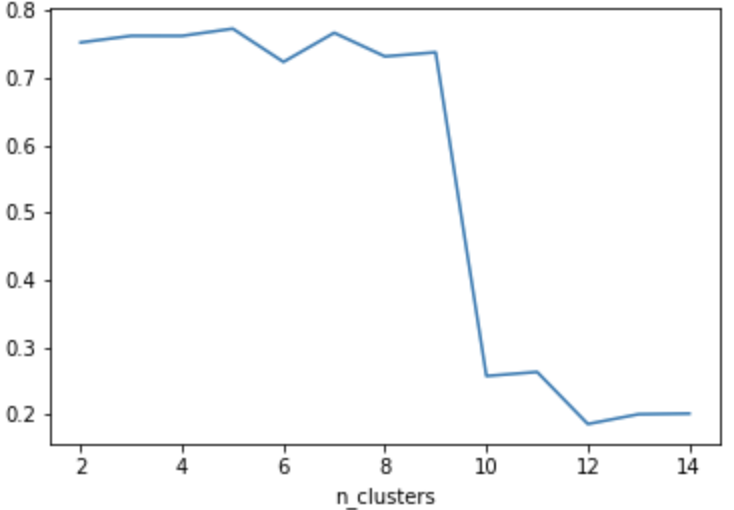}  
  \caption{Silhouette score suggest 9 groups.}
  \label{per_siluet}
\end{subfigure} \hfill
\begin{subfigure}[t]{.275\textwidth}
  \centering
  \includegraphics[width=\linewidth]{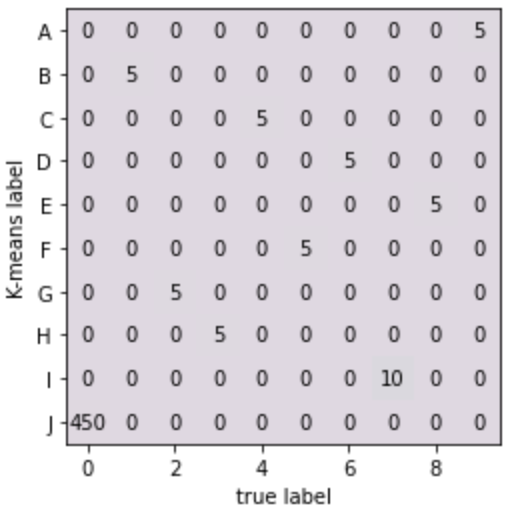}  
  \caption{Confusion matrix for 10 groups. All observations of a same group ($x$-axis) are assigned to the same cluster ($y$-axis).}
  \label{per_confmat_10}
\end{subfigure} \hfill
\begin{subfigure}[t]{.3\textwidth} 
  \centering
  \includegraphics[width=\linewidth]{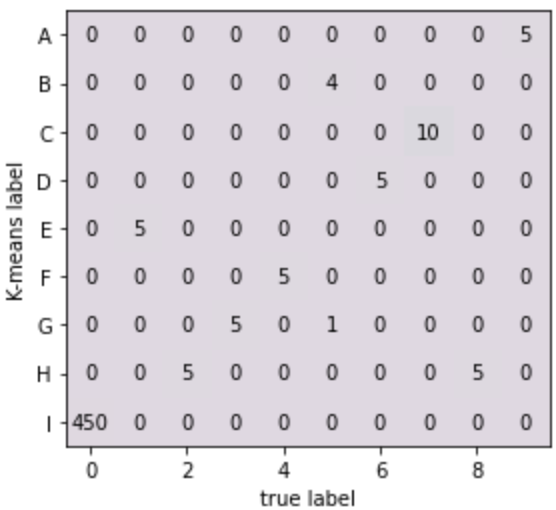}  
  \caption{Confusion matrix for 9 groups. We can see that anomaly types $2$ and $8$ (Figure \ref{data_per}) are merged in cluster $H$. It can be explained by the fact that the parameter $X_1$ contains the same type of anomalous behavior (white noise during whole observation) in groups $2$ and $8$.}
  \label{per_confmat_9}
\end{subfigure}
\caption{Clustering in \it{Dataset 1}}
\label{IF groups D1}
\end{figure}

\begin{figure}[!ht]
\caption{Overview of features importance per group found by SHAP.} 
\centering
\includegraphics[width=0.7\textwidth]{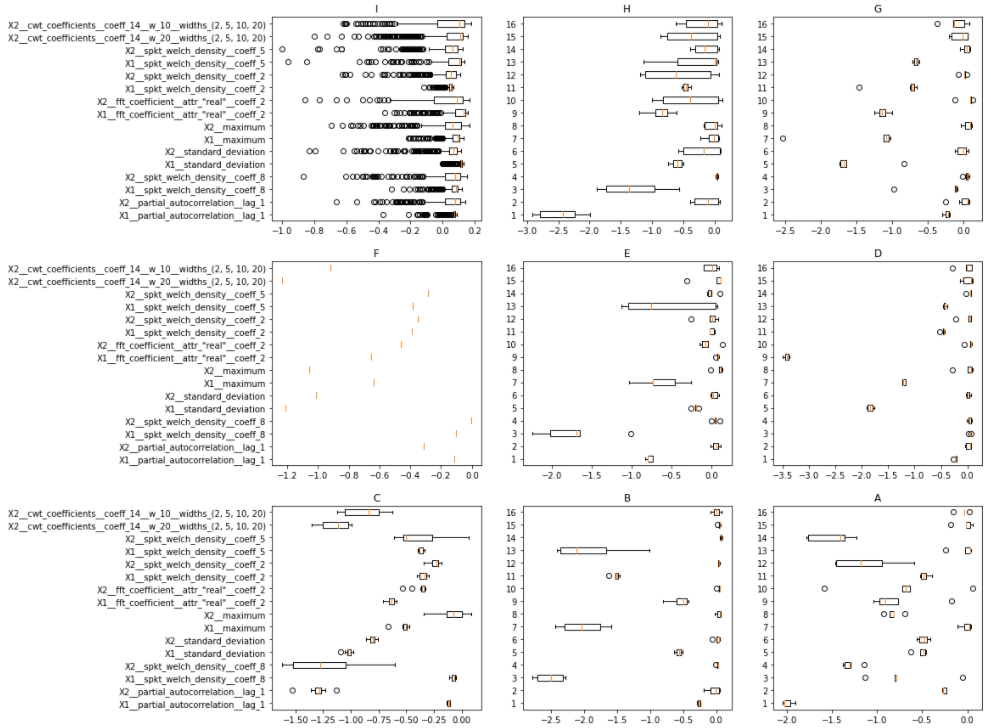}
\label{per_shap}
\end{figure}

\begin{figure}[!ht]
\caption{Overview of features importance ($y$-axis) found by a decision tree. The criterion is Gini importance ($x$-axis), i.e. reduction of the split impurity brought by a feature.} 
\centering
\includegraphics[width=0.9\textwidth]{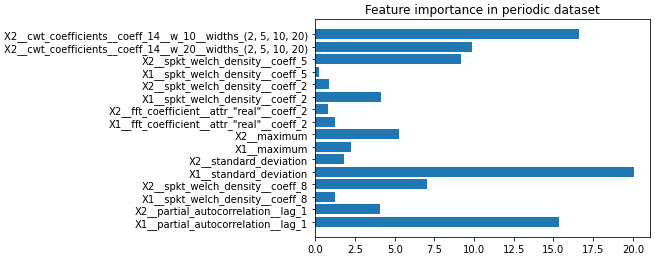}
\label{per_tree_features_shap}
\end{figure}

We use the silhouette score to determine the number of groups in {\it Dataset 1}. Figure \ref{per_siluet} suggests 9 groups. Although the clustering on 10 groups was correct (see the confusion matrix in Figure \ref{per_confmat_10}), we will proceed with 9 groups as we want to reproduce a situation when we do not know the exact number of anomaly groups in advance. Features importance is presented in Figures \ref{per_shap} and \ref{per_tree_features_shap}.

\subsection{Real data: Model and evaluation}
\label{Real Model and evaluation}

\begin{figure}[!ht]
\caption{Low score observations in \it{Dataset 3}. Five series of $X_2$ are displayed. Our method does not consider slight oscillations as anomalies. It could be possible to target the detection of this phenomena by choosing a suitable feature. However, a client  has confirmed that this phenomena is not considered as an anomaly.} 
\centering
\includegraphics[width=.6\textwidth]{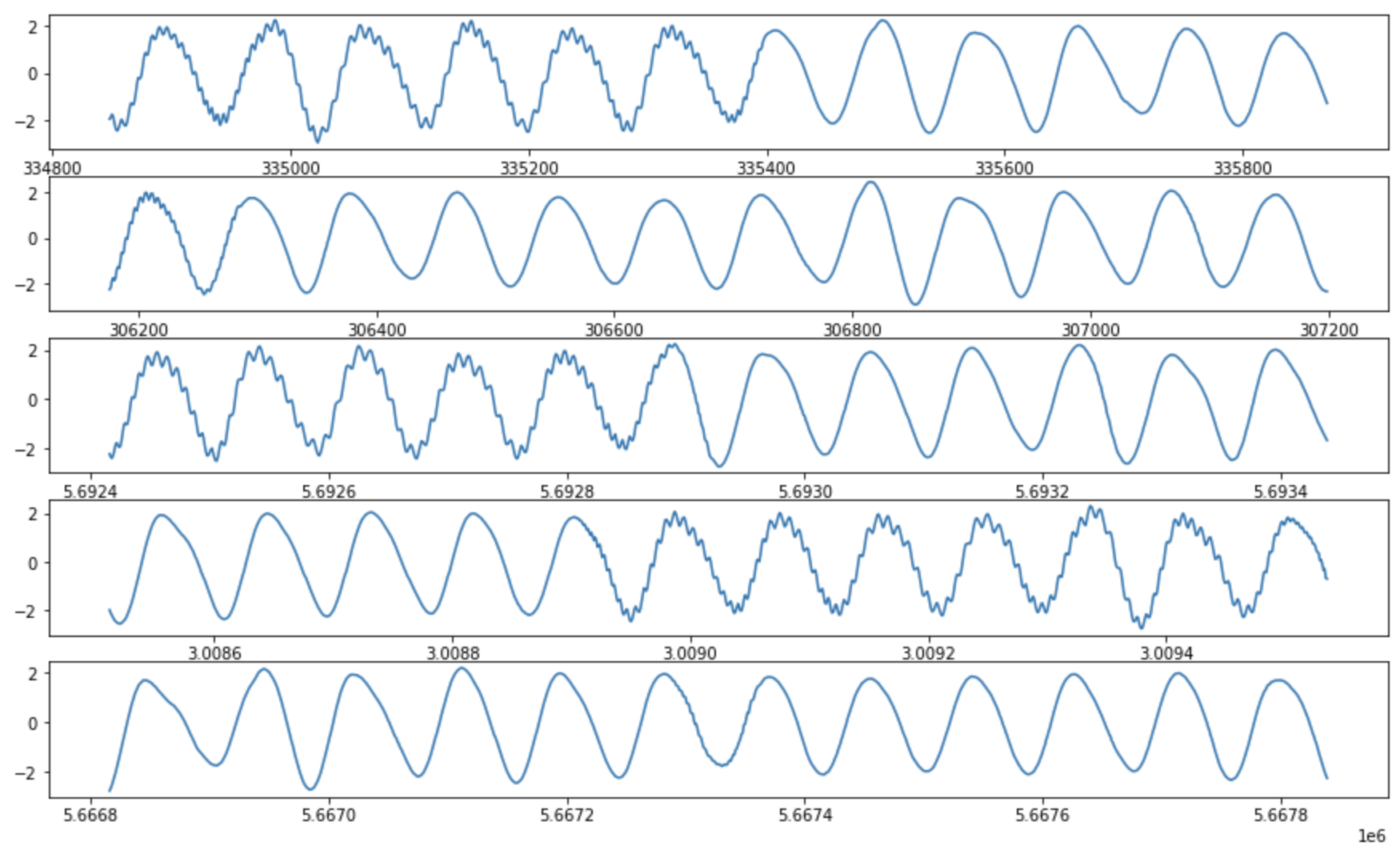}
\label{real_normal}
\end{figure}

\begin{figure}[!ht]
\caption{High score observations in \it{Dataset 3}. Five series of $X_2$ are displayed. Indeed, we can see unusual fluctuations in those series.} 
\centering
\includegraphics[width=0.6\textwidth]{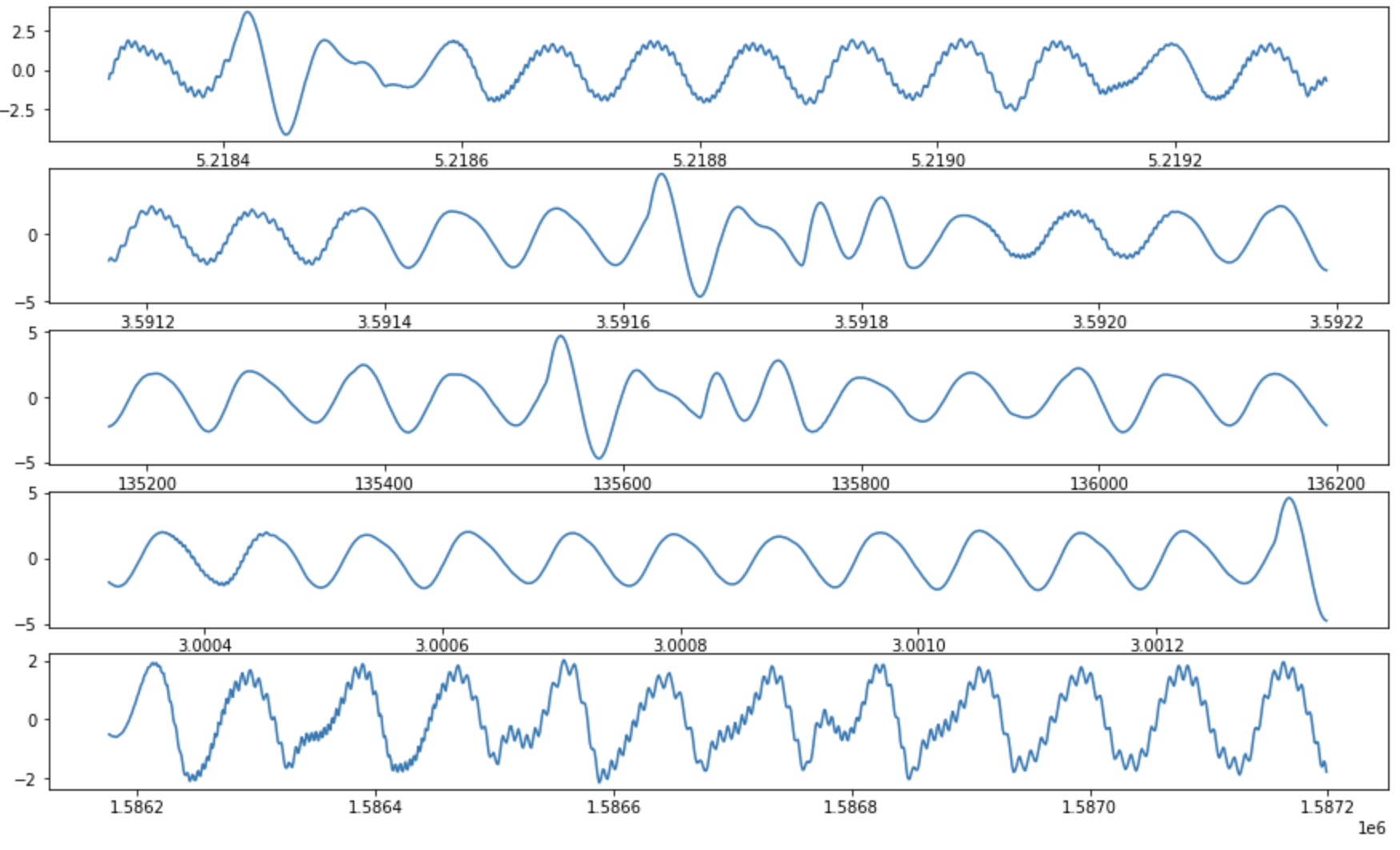}
\label{real_anormal}
\end{figure}

As with generated datasets, the input of the Isolation forest algorithm is based on the vectors of extracted features. Since we do not know exactly which observations are anomalous, we visually compare selected low anomaly score observations with selected high anomaly score observations (see Figures \ref{real_normal} and \ref{real_anormal}).

\begin{figure}[!ht]
\begin{subfigure}[t]{.5\textwidth} 
  \centering
  \includegraphics[width=1.0\linewidth]{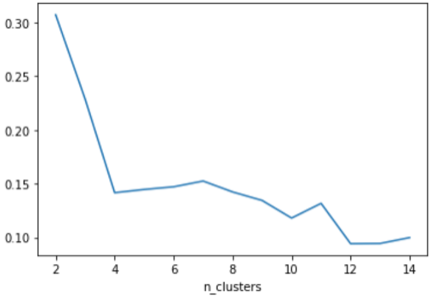}  
  \caption{Silhouette score is used to suggest the possible number of clusters in the dataset.}
  \label{siluet_real_data}
\end{subfigure}
\begin{subfigure}[t]{.5\textwidth}
  \centering
  \includegraphics[width=1.0\linewidth]{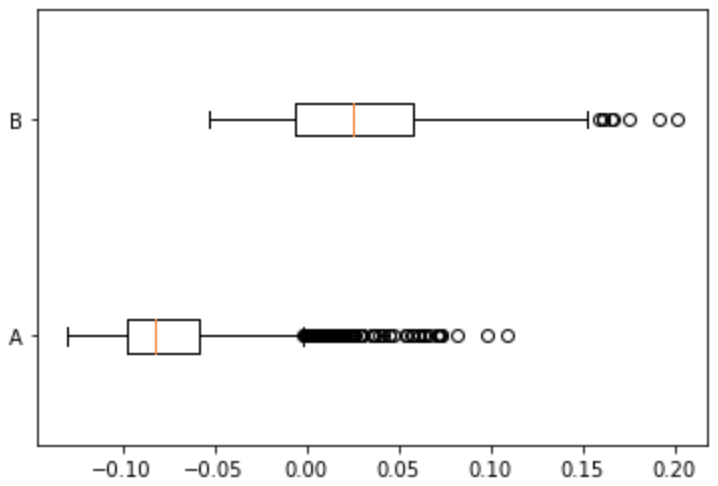}  
  \caption{Comparison of the Isolation forest anomaly score per group found with K-means. Group A has visibly lower anomaly score than group B. Hence we can assume that group A contains mostly normal observations and group B contains mostly anomalies.}
  \label{real_IF}
\end{subfigure}
\caption{Examination of possible anomaly groups in \it{Dataset 3}}
\label{IF groups r}
\end{figure}

\begin{figure}[!ht]
\centering
\includegraphics[width=1\textwidth]{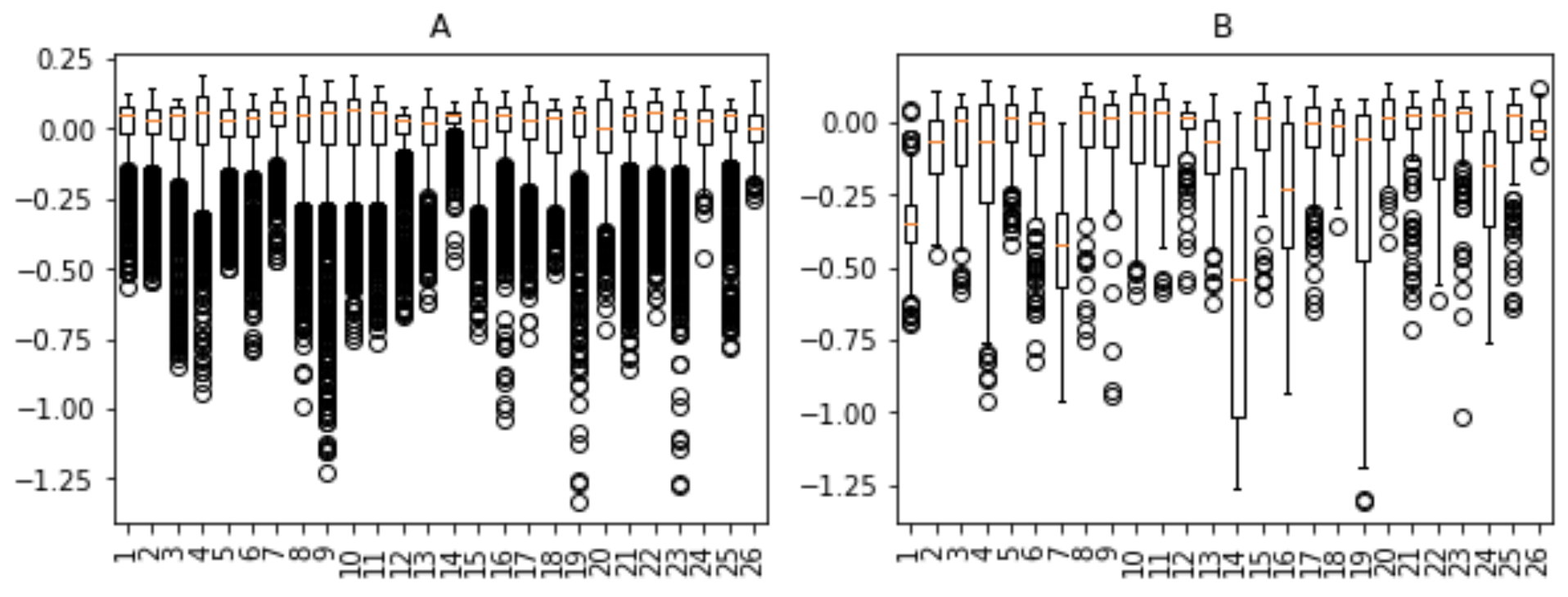}
\caption{ SHAP importances of the features by group. For simplicity, features are encoded. 
} \label{real_shap}
\end{figure}

\begin{figure}[!ht]
\centering
\includegraphics[width=.5\textwidth]{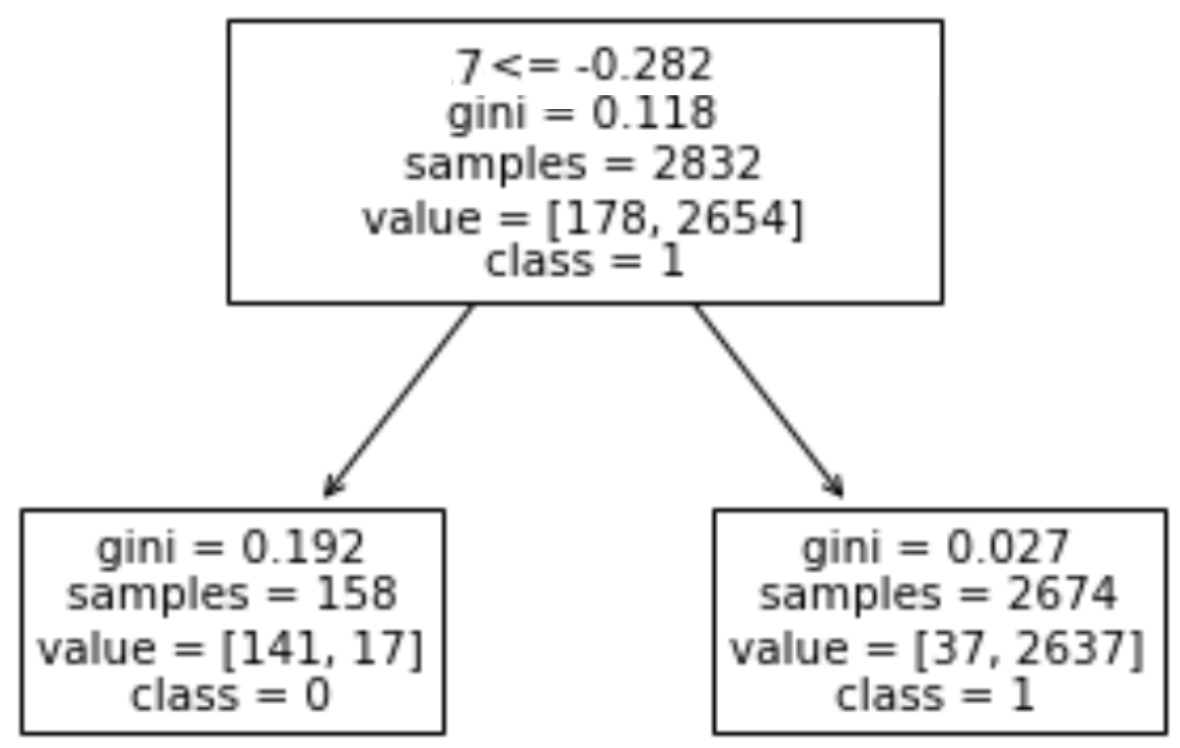}
\caption{ Supervised decision tree trained on the K-means labels and SHAP values. Class $1$ correspond to cluster A, while class $0$ corresponds to cluster B (see Figures \ref{IF groups r} and \ref{real_shap}). Only one feature (feature \texttt{7 = minimum of $X_2$}) is sufficient to create the clusters. Although there are impurities in this split, we can assume that this feature - minimal value of the second parameter - describes many of the real anomalies, which is visually confirmed in Figure \ref{real_anormal}.
} \label{real_tree}
\end{figure}

It is not known whether there are more anomaly types present in the dataset. The suitable number of clusters is fixed by using the silhouette score. The analysis of the detected groups are in Figures \ref{IF groups r}, \ref{real_shap} and \ref{real_tree}.

\section{Conclusion}
\label{Conclusion}

This paper has presented a procedure to detect multi-class anomalies and to identify the features describing these anomalies in case of a functional dataset.
This approach is flexible and not restricted to data following a specific distribution. It provides good results, detecting various kind of anomalies in multivariate datasets while each step in the process remains fast and comprehensible.
The most challenging part lies in choosing the features. By now the process is individualized for each dataset but a possible future work would be to choose default groups of the features depending on the character of the data without great loss of information.
An other possible work would be to improve the clustering algorithm and metrics to automatically fix the number of clusters.
The method can be extended to other types of data. More research is needed to separate a continuous non-repetitive process into homogeneous segments and extract relevant features but the rest of the process (anomaly detection and their explainability) could be applied in the same way.

\bibliographystyle{alpha}
\bibliography{articleRDJS2.bib}

\end{document}